\documentclass[letterpaper, 10 pt, conference]{ieeeconf}  

\usepackage{graphicx}
\usepackage{amsmath}
\usepackage{amssymb}
\usepackage{booktabs}
\usepackage{hyperref}
\hypersetup{colorlinks}
\usepackage{todonotes}
\usepackage{xcolor}
\usepackage{caption}
\usepackage{subcaption}
\usepackage{mathtools}
\usepackage{svg}
\usepackage{stackengine}
\usepackage{siunitx}
\newboolean{long_appendix}
\setboolean{long_appendix}{false}

\usepackage{todonotes}

\usepackage[super]{nth}
\usepackage[capitalize]{cleveref}
\crefname{section}{Sec.}{Secs.}
\Crefname{section}{Section}{Sections}
\Crefname{table}{Table}{Tables}
\crefname{table}{Tab.}{Tabs.}

\usepackage{balance}

\IEEEoverridecommandlockouts                              

\usepackage{newtxtext} 
\usepackage{newtxmath} 
\usepackage{microtype}
\usepackage{times} 

\title{\LARGE \bf
\emph{PEGASUS}: Physically Enhanced Gaussian Splatting Simulation System for 6DoF Object Pose Dataset Generation
}

\author{Lukas Meyer\textsuperscript{1,\textdagger}, Floris Erich\textsuperscript{2}, Yusuke Yoshiyasu\textsuperscript{2}, Marc Stamminger\textsuperscript{1}, Noriaki Ando\textsuperscript{2} and Yukiyasu Domae\textsuperscript{2}%
\thanks{$^{1}$Visual Computing Erlangen, Friedrich-Alexander-Universität Erlangen-Nürnberg-Fürth, Germany
        {\tt\small lukas.meyer@fau.de}}%
\thanks{$^{2}$Industrial CPS Research Center, National Institute of Advanced Industrial Science and Technology, Japan
        }%
\thanks{\textdagger This work was conducted during an internship at the Industrial CPS Research Center, National Institute of Advanced Industrial Science and Technology. This paper is one of the achievements of joint research with and is jointly owned copyrighted material of ROBOT Industrial Basic Technology Collaborative Innovation Partnership. This research has been supported by the New Energy and Industrial Technology Development Organization (NEDO), under the project ID JPNP20016.}%
}

\begin{document}

\twocolumn[{
\renewcommand\twocolumn[1][]{#1}
\maketitle

\includegraphics[width=\textwidth]{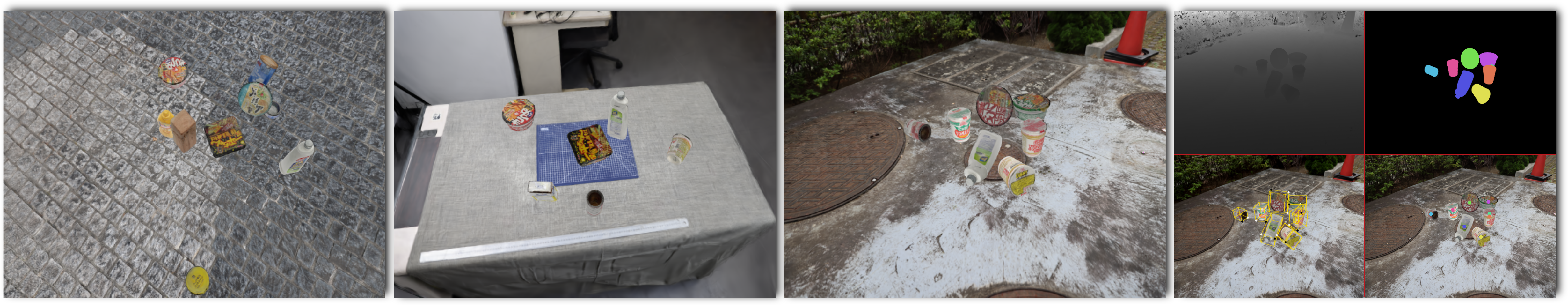}

\captionof{figure}{Representative scenes generated by \emph{PEGASUS}. By separately reconstructing objects and environment with Gaussian Splatting and connecting them to a physics engine a vast variety of scenes can be generated utilizing novel view synthesis. At each snapshot, multiple data points such as RGB images, segmentation masks, depth maps, 2D/3D bounding boxes, and object poses can be extracted.} 
\vspace{1em}
\label{fig:tile_figure}
}]

\thispagestyle{empty}
\pagestyle{empty}

\footnotetext[1]{Visual Computing Erlangen, Friedrich-Alexander-Universität Erlangen-Nürnberg-Fürth, Germany. E-Mail:
        {\tt\small lukas.meyer@fau.de}}
\footnotetext[2]{National Institute of Advanced Industrial Science and Technology, Tokyo, Japan}
\renewcommand{\thefootnote}{\fnsymbol{footnote}}
\footnotetext[2]{This work was conducted during an internship at the National Institute of Advanced Industrial Science and Technology.} 
\footnotetext[1]{This paper is one of the achievements of joint research with and is owned copyrighted material of ROBOT Industrial Basic Technology Collaborative Innovation Partnership. This research has been supported by the New Energy and Industrial Technology Development Organization (NEDO), under the project ID JPNP20016.} 
\renewcommand{\thefootnote}{\arabic{footnote}}
\setcounter{footnote}{0}

\begin{abstract}
We introduce \underline{P}hysically \underline{E}nhanced \underline{Ga}ussian \underline{S}platting Sim\underline{u}lation \underline{S}ystem (\emph{PEGASUS}) for 6DoF object pose dataset generation, a versatile dataset generator based on 3D Gaussian Splatting. 
Environment and object representations can be easily obtained using commodity cameras to reconstruct with Gaussian Splatting.
\emph{PEGASUS} allows the composition of new scenes by merging the respective underlying Gaussian Splatting point cloud of an environment with one or multiple objects. 
Leveraging a physics engine enables the simulation of natural object placement within a scene through interaction between meshes extracted for the objects and the environment. 
Consequently, an extensive amount of new scenes - static or dynamic - can be created by combining different environments and objects. 
By rendering scenes from various perspectives, diverse data points such as RGB images, depth maps, semantic masks, and 6DoF object poses can be extracted.
Our study demonstrates that training on data generated by \emph{PEGASUS} enables pose estimation networks to successfully transfer from synthetic data to real-world data.
Moreover, we introduce the \emph{Ramen} dataset, comprising 30 Japanese cup noodle items. 
This dataset includes spherical scans that capture images from both the object hemisphere and the Gaussian Splatting reconstruction, making them compatible with \emph{PEGASUS}.

\end{abstract}

\section{Introduction}

Robotic manipulation in changing environments with various novel objects pose significant challenges. 
Tasks such as object detection, segmentation, and pose estimation require substantial training data to adapt to new scenarios.
The creation of suitable datasets can be achieved either by annotating real-world data (e.g., YCB-V~\cite{PoseCNN}, HOPE~\cite{HOPE}, LINEMOD~\cite{LINEMOD}) or by utilizing synthetic datasets (e.g., Fallen Things (FAT)~\cite{FAT}).
NDDS~\cite{NDDS} and BlenderProc~\cite{blenderproc2} synthetically generate domain-specific datasets, facilitating the straightforward insertion of modeled object assets into synthetic environments. 
However, training on synthetically generated datasets may be compromised by a lack of realism and the time-consuming process of creating detailed models for specific environments and objects.

To address these limitations and reduce the reality gap, we make use of advanced novel view synthesis techniques. 
The rendering quality of novel view synthesis methods, such as Neural Radiance Fields (NeRF)~\cite{nerf} have steadily increased over the past years \cite{nerf_survey}.
With 3D Gaussian Splatting (3DGS)~\cite{GaussianSplatting}, a real-time method using explicit representation, it becomes possible to create more realistic scenes. By leveraging this explicit representation, we can build a modular pipeline by combining multiple Gaussian splatting point clouds for dataset generation. 
This approach allows for the rapid and flexible generation of high-quality, realistic synthetic data, which can significantly enhance training and adaptation for robotic manipulation tasks in diverse environments.

In this work, we introduce \emph{PEGASUS}, a physically enhanced Gaussian Splatting simulation environment, designed to create innovative datasets for 6DoF object pose estimation.
By utilizing 3DGS, it allows us to close or at least narrow the gap between synthetic training data and real-world applications.
Furthermore, it is very simple to generate custom assets: new models are obtained by scanning real-world objects and environments and reconstructing them in an automated pipeline.
By combining 3DGS assets a comprehensive dataset can be created to fine-tune object pose estimation networks for desired operating environments.

In \emph{PEGASUS}, we separately create the environment and objects with Gaussian Splatting and simulate the interaction of both elements using a physics engine. 
In this matter, we can render novel views for RGB images, semantic masks, depth maps, and metadata such as the object pose and  2D/3D bounding boxes. 
By extracting the data in the \textit{Benchmark for 6D Object Pose Estimation} (BOP) data format~\cite{BOP} it can be easily used to train pose estimation networks and other network types. 

Our experiments demonstrate that the pose estimation network, Deep Object Pose (DOPE)~\cite{DOPE} when trained on \emph{PEGASUS}-generated dataset, can operate a grasping task with Universal Robots \textit{UR5} based on our dataset and successfully shows synthetic to real transfer. In summary, we
make the following contributions:
\begin{itemize}
    \item \emph{PEGASUS}: A dataset generation tool for photo-realistic 6DoF object pose estimation, utilizing 3D Gaussian Splatting. The tool's code has been made open-source\footnote{\emph{PEGASUS Code}: \url{https://github.com/meyerls/PEGASUS}}.
    \item \emph{Ramen} Dataset\footnote{\emph{Ramen}-Dataset: \url{https://meyerls.github.io/pegasus_web}}: A comprehensive collection of over 30 products, featuring images, COLMAP reconstructions, and 3D Gaussian Splatting reconstructions.
    \item \emph{PEGASET} Dataset: A collection of scanned environments and 21 re-scanned objects from YCB-V.
\end{itemize}

\section{Related Work}

\subsubsection*{\textbf{Neural Radiance Fields}}

Novel view synthesis has recently become a popular research topic that studies techniques for generating novel views of captured scenes. 
Various approaches, including implicit representations, voxels, and point clouds, are used for scene representation.

Neural Radiance Fields (NeRF)~\cite{nerf} build a continuous implicit representation by optimizing a Multi-Layer Perceptron (MLP) through volumetric rendering. 
This process encodes information within the MLP weights, requiring network queries for every spatial point to extract color and density data.
Consequently, editing NeRFs involves retraining, which is computationally expensive.

InstantNGP \cite{instantNGP} employs a multi-resolution hash grid for spatial information storage, resulting in much faster access and thus training and rendering time.
However, scene modifications necessitate altering the grid structure, as demonstrated by NeRFShop~\cite{NeRFShop}.
NeRFShop enables volumetric manipulation, yet this involves interactive region selection and manual object deformation, and it is not conducive to automation.
CLIP-NeRF~\cite{clipnerf} integrates CLIP~\cite{CLIP} to manipulate the shape and appearance of NeRF by training a deformation network, which limits its suitability for rapid editing.

Alternatives are non-volumetric representations, based on surfaces or points.
Such representations are not trained from scratch, but start with a point cloud reconstructed using standard structure-from-motion methods \cite{schoenberger2016sfm}, SLAM or LIDAR.
ADOP~\cite{ADOP} and TRIPS \cite{franke2024trips}  fall into this class, they utilize point cloud-based radiance fields, enhancing points with neural features. 
These approaches achieve high rendering quality and fast rendering time, but the optimization of the neural features and the rendering network is still time-consuming.

3D Gaussian Splatting (3DGS)~\cite{GaussianSplatting} also starts with a reconstructed point cloud, but renders these as semi-transparent Gaussian splats.
Such splats are a powerful rendering primitive, which makes it possible to optimize them directly, without requiring neural parameters and a final neural rendering network.
3DGS show high rendering quality at a very high rendering speed.
The specialized differential Gaussian rasterization pipeline facilitates straightforward manipulation (transformation, insertion, deletion) of the underlying point cloud. 
This enables an easy and rapid combination of different reconstructions, a crucial aspect of \emph{PEGASUS}, which is why we 3DGS as the basis for our approach.

\subsubsection*{\textbf{Dataset Generation}}

Existing datasets are typically categorized as either synthetic or real-world. 
Synthetic datasets, like BlenderProc~\cite{blenderproc2} or NVIDIA Deep Learning Dataset Synthesizer (NDDS)~\cite{NDDS}, offer the advantage of generating numerous unique scenes. 
However, they face challenges in asset modeling and achieving photorealistic rendering, often resulting in a domain gap when applied to real-world scenarios. 
To mitigate this, physically-based rendering techniques are employed, incorporating complex lighting effects such as scattering, refraction, and reflection, to enhance realism~\cite{hodan2019photorealistic}.

Conversely, creating real-world datasets, such as YCB-V \cite{PoseCNN}, is a labor-intensive process that requires meticulous annotation, often prone to human error. 
Capturing a wide variety of scenes to ensure dataset variance further adds to the complexity. 
Despite these challenges, real-world datasets generally offer better generalization for deep learning applications than their synthetic counterparts.

NeuralLabeling~\cite{neurallabeling} offers to directly annotate Neural Radiance fields and precisely extract the underlying object structures. To create a large dataset it is still time-consuming to generate a vast variety of scenes.

\emph{PEGASUS} adopts a hybrid approach, combining the strengths of both synthetic and real-world datasets. 
Leveraging the modularity of synthetic data, it allows for the generation of new scenes through the combination of scanned objects and diverse environments, leading to a multitude of data points. 
Additionally, \emph{PEGASUS} employs novel view synthesis techniques to render photorealistic scenes that are practically indistinguishable from real-world data.

\begin{figure*}[ht!]
    \centering
    \includegraphics[width=1\linewidth]{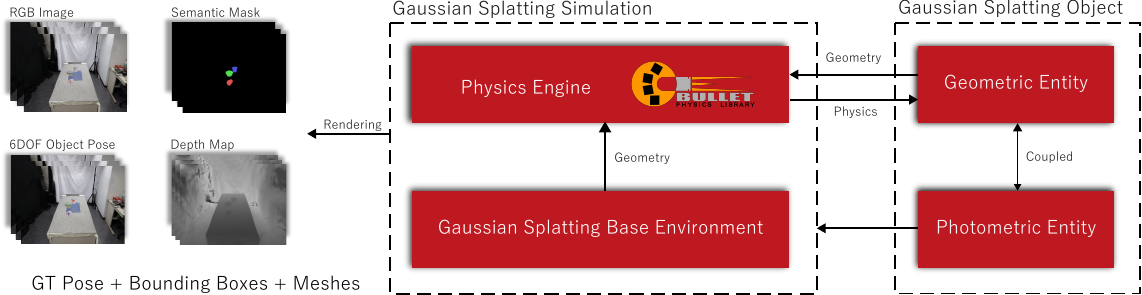}
    \caption{Pipeline of the \emph{PEGASUS} dataset generator. The 3DGS base environment (see Section \ref{subsec:base_environment}) comprises both the 3DGS reconstruction and a mesh reconstructed from its point cloud. The 'Object' includes the 3DGS representation of the object (discussed as the photometric entity in Section \ref{subsec:photometric_entity}) and a low-poly mesh of the same object (covered as the geometric entity in Section \ref{subsec:geometric_entity}). By utilizing the mesh of the base environment and the object entity, an arbitrary number of objects can be simulated in the physics engine (refer to Section \ref{subsec:physical_engine}), facilitating realistic and random placement of the objects within the scene.
    When the trajectories of the objects are applied to the photometric instances of the environment and the object, we are capable of rendering dynamic and static scenes from various viewpoints and time steps. These data are then saved in the BOP data format \cite{BOP}.}
    \label{fig:pegasus_pipeline}
        \vspace{-0.2cm}

\end{figure*}

\section{Prerequisites}

\subsection{Gaussian Splatting}
\label{subsec:gaussian_splatting}

3D Gaussian Splatting \cite{GaussianSplatting} is an efficient method for performing novel view synthesis from captured scenes. It utilizes an unstructured, discrete representation in the form of a point cloud, which offers significant flexibility for modifying and manipulating inherent geometry.

The input for Gaussian Splatting comprises a set of images capturing a static scene or object, corresponding poses, camera intrinsics, and a sparse point cloud, which are typically generated using Structure from Motion (SfM)~\cite{schoenberger2016sfm}. 
This sparse point cloud is then transformed into a more complex 3D Gaussian Splatting point cloud, denoted as $\mathbf{P}_{GS} = \{\mathbf{P}_{\mathbf{\mu}}, \mathbf{P}_\Sigma, \mathbf{P}_\alpha, \mathbf{P}_{\mathbf{f}}\}$. Each point in this cloud, represented by $\mathbf{x}$ (also interpretable as the mean position $\mathbf{\mu}$ of the Gaussian), is associated with a covariance matrix $\mathbf{\Sigma}$, an opacity value $\alpha$, and a set of spherical harmonic coefficients $\mathbf{f}$, which are used for directional appearance coloring.

The 3D Gaussians have to be projected onto the 2D image plane to optimize the parameters of the Gaussian Splatting point cloud. Therefore a differential tile-based Gaussian rasterization pipeline proposed by \cite{GaussianSplatting} is utilized. Each Gaussian is characterized by
\begin{equation}
	G(\mathbf{x})~= e^{-\frac{1}{2}(\mathbf{x})^{T}\mathbf{\Sigma}^{-1}(\mathbf{x})}
\end{equation}
where $\mathbf{\Sigma}$ is a full 3D covariance matrix defined in world space and centered at the point means $\mathbf{\mu}$. 
By projecting the 3D Gaussians back onto the 2D image the covariance matrix in image space \cite{EWA} is computed through:
\begin{equation}
	\mathbf{\Sigma'} = \mathbf{J} \mathbf{W} \mathbf{\Sigma} \mathbf{W}^{T} \mathbf{J}^{T}.
\end{equation}
$\mathbf{W}$ is defined as the transformation matrix from world to camera space and $\mathbf{J}$ is the Jacobian of the affine approximation of the projective transformation. 

After mapping the 3D Gaussians onto the 2D image plane the alpha-blended rendering is performed
for each pixel in front-to-back depth order to evaluate the final color and alpha values~\cite{GaussianSplatting}. The blending of the $N$ ordered sample points in a pixel is computed by:
\begin{equation}
	\label{eq:front-to-back}
	\mathbf{C} = \sum_{i \in N}
	\mathbf{c}_{i}\alpha_{i}
	\prod_{j=1}^{i-1}(1-\alpha_{j}).
\end{equation}
$\mathbf{c_i}$ and $\alpha_i$ are defined as color and opacity of the $i$th Gaussian. 

\subsection{6-DOF Manipulation of Gaussian Splatting}
\label{subsec:manipulation}

The manipulation of Gaussian Splatting benefits from its underlying explicit representation. 
The appearance of the Gaussian point cloud $\mathbf{P}_{GS}$ is defined by the Gaussian's $\mathbf{P}_{\Sigma}$, 3D mean value $\mathbf{P}_{\mu}$, opacity values $\mathbf{P}_{\alpha}$ and coefficients $\mathbf{P}_{f}$ of the spherical harmonics. 
For manipulating a Gaussian point cloud only   $\mathbf{P}_{\mu}$, $\mathbf{P}_{\Sigma}$ and $\mathbf{P}_{f}$ are relevant as the scalar opacity values in $\mathbf{P}_{\alpha}$ do not change if a transformation is applied. 
For applying a transformation matrix $\mathbf{T} = [ \mathbf{R} | \mathbf{t} ]$ the translational and rotational part has to be considered separately.

The translational part $\mathbf{t}$ is a straightforward operation. 
This involves applying a translation vector $\mathbf{t}_{\Delta} = (x_{\Delta}, y_{\Delta}, z_{\Delta})^\top$ to the mean values $\mathbf{x} = (x, y, z)^\top$ of the Gaussians. 
The remaining parts of the Gaussian point cloud (such as $\mathbf{P}_{\Sigma}$ and $\mathbf{P}_{f}$) are not affected.

Rotating a Gaussian Splatting point cloud is more complex. 
Here, a rotation matrix $\mathbf{R}$ has to be applied on the points $\mathbf{P}_{\mu}$, the covariance matrices $\mathbf{P}_{\Sigma}$ and the coefficients $\mathbf{P}_{f}$ of the spherical harmonics.
For $\mathbf{P}_{\mu}$ and $\mathbf{P}_{\Sigma}$ the rotation is directly applied to their corresponding point clouds.

To also obtain the identical view-dependent effects as for the original scene one has to apply a rotation also to the spherical harmonics \cite{GreenSH}.
This is commonly done by not rotating the base functions of the spherical harmonics but rather their coefficients \cite{GreenSH}. 
In \cite{Kautz} a method is proposed to decompose the rotation matrix $\mathbf{R}$ into its Euler angles and build a block diagonal sparse matrix \cite{GreenSH} to rotate every band individually.
Due to the rotational in-variance of spherical harmonics, this rotation is lossless.
For a detailed explanation please refer to \cite{Kautz, GreenSH}.

\section{Methodology}

The core principle of \emph{PEGASUS} is the separate consideration of the environment and individual objects. 
By situating a set of objects within multiple environments, a vast set of scenes can be created. 
The integration of the physics engine \emph{PyBullet}~\cite{pybullet} into \emph{PEGASUS} enables the simulation of natural object placement in scenes and the creation of dynamic scenes within the Gaussian Splatting Simulation Environment.

\subsection{Gaussian Splatting Simulation Environment}
\label{subsec:gs_simulation}

The core pipeline of \emph{PEGASUS}, as illustrated in Fig. \ref{fig:pegasus_pipeline}, is composed of several distinct blocks. Below, we detail the essential components of this pipeline: the base environment, the creation of Gaussian Splatting objects, and the integration of a physics engine into our simulation setup.
Together, these blocks form the backbone of the \emph{PEGASUS} pipeline, enabling the creation of complex, multi-modal datasets that closely mimic real-world conditions.

\subsubsection{Base Environment}
\label{subsec:base_environment}

The base environment serves as the foundational construct for building the actual scene in \emph{PEGASUS}. 
To create this, we recorded 9 different planar scenes using a DSLR camera, capturing between 100 and 150 images per scene. 
We then used SfM to recover the set of poses and a sparse point cloud $ \mathbf{P}_{\text{sparse}}$.

The scenes are then automatically scaled to obtain a true-to-scale metric reconstruction by placing an ArUco marker in the scene ~\cite{CherryPicker} and afterward aligning the planar area to the center so that the $\mathbf{z}$ vector points upwards.

Subsequently, we performed Gaussian Splatting on the scene using the default parameters suggested by 3DGS~\cite{GaussianSplatting}. 
Towards the end of this process, we extracted the plain point cloud by obtaining the mean value of each Gaussian splat. To integrate with the physics engine, we converted this point cloud into a mesh. For this purpose, we employed the alpha shape algorithm \cite{edelsbrunner1994threedimensional} as a mesh recovery technique, transforming the 3DGS-extracted points into a geometric mesh.

\begin{figure*}[htp]
    \centering
    \includegraphics[width=0.9\linewidth]{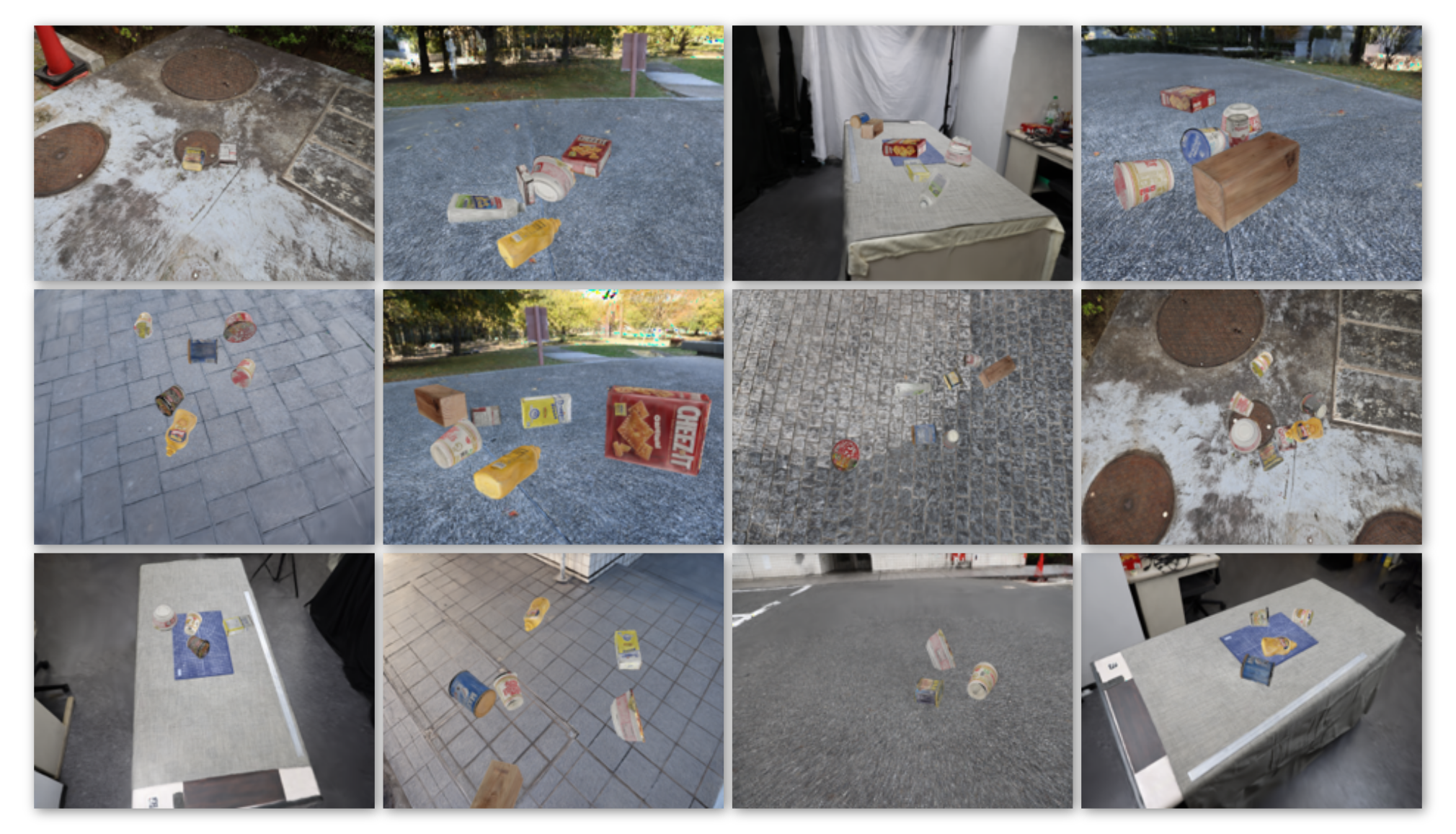}
    \caption{Gallery of data generated by \emph{PEGASUS}. It shows scenes generated with 9 different base environments and an arbitrary combination of the 30 elements from the \emph{Ramen} dataset and from the 21 YCB objects \cite{YCB_Objects} from the YCB-V dataset.}
    \label{fig:gallary_images}
    \vspace{-0.2cm}
\end{figure*}

\subsubsection{Gaussian Splatting Object}
\label{subsec:gs_object}

The reconstruction of objects in \emph{PEGASUS} follows a procedure similar to that of the base environment.
We utilize an Ortery scanning system~\cite{Ortery3DPhotoBench280} for image acquisition. 
Detailed information about this process and the various types of objects are discussed in Section~\ref{subsec:konbini_dataset}. 
Below, we introduce the concept of the Gaussian Splatting object, which comprises two main components: the photometric and geometric entities.

\subsubsection*{Photometric Entity}
\label{subsec:photometric_entity}

The photometric entity is crucial for rendering objects using Gaussian Splatting. 
We start with a spherical, sparse, and metrically reconstructed Structure from Motion (SfM) model of the object as input. 
Using Gaussian Splatting with the same settings as in the base environment setup, we generate this photometric entity. It can then be integrated into the simulation, enabling simultaneous rendering of both the base environment and the object.

\subsubsection*{Geometric Entity}
\label{subsec:geometric_entity} 

For the geometric entity, we start with the colored point cloud extracted from the Gaussian Splatting reconstruction. 
The point cloud is initially cleaned by removing outliers, and then the alpha shape algorithm \cite{edelsbrunner1994threedimensional} is applied for mesh reconstruction. 
To achieve a smoother surface, we use Laplacian smoothing \cite{nealen2006laplacian}. 
The geometric entity is stored as a low-polygon triangle mesh, optimizing computational efficiency when simulating this mesh with the physics engine.

\subsubsection{Physics Engine}
\label{subsec:physical_engine}

We selected PyBullet~\cite{pybullet} as our lightweight physics engine. 
Within the simulation, the environment mesh is integrated as a static component. 
For the objects, we determine an appropriate height to drop varying quantities (user-selected) into the scene, simulating natural object placement. 
Throughout the simulation, we track and extract the orientation and translation of each object, represented as a quaternion and a translation vector, respectively. 
\emph{PEGASUS} is thus equipped to simulate both static and dynamic scenes, offering a comprehensive range of possibilities for scene creation and analysis.

\subsection{\emph{PEGASUS} Dataset Generation}
\label{subsec:pegasus_dataset_generator}

To create a dataset using Gaussian Splatting with \emph{PEGASUS}, we include Gaussian Splatting base environments and Gaussian Splatting objects. 
This approach, enhanced by physical placement techniques, allows us to generate an unlimited number of unique scenes. 
To generate a new scene, we simulate the trajectory using our physics engine, apply transformations to each Gaussian Splatting object, and render the scene with the Gaussian Splatting rasterizer~\cite{GaussianSplatting}.

To render the scene from various viewpoints, we randomly select a set of ground truth poses and create a trajectory by interpolating between these poses.
Lastly, we save the raw data, including rendered RGB images, depth maps, segmentation maps of the silhouette and visibility masks, 2D/3D bounding boxes as well as the transformation matrices from object to world and world to camera coordinate systems. 
This data is formatted according to the popular BOP-dataset format~\cite{BOP}.
We want to emphasize that \emph{PEGASUS} is also capable of rendering dynamic scenes, which broadens the potential applications of the generated data. 
It is worth noting that this approach can be easily extended with custom data. 
To incorporate a new environment, one only needs to record a set of images and convert them into a 3DGS instance. 
The same process applies to any object; a simple scan is enough to compute the photometric and geometric properties required for integration into \emph{PEGASUS}. A set of images extracted from the dataset generator is shown in Fig. \ref{fig:gallary_images}.

\begin{figure}[b!]
	\centering
	\includegraphics[width = \linewidth]{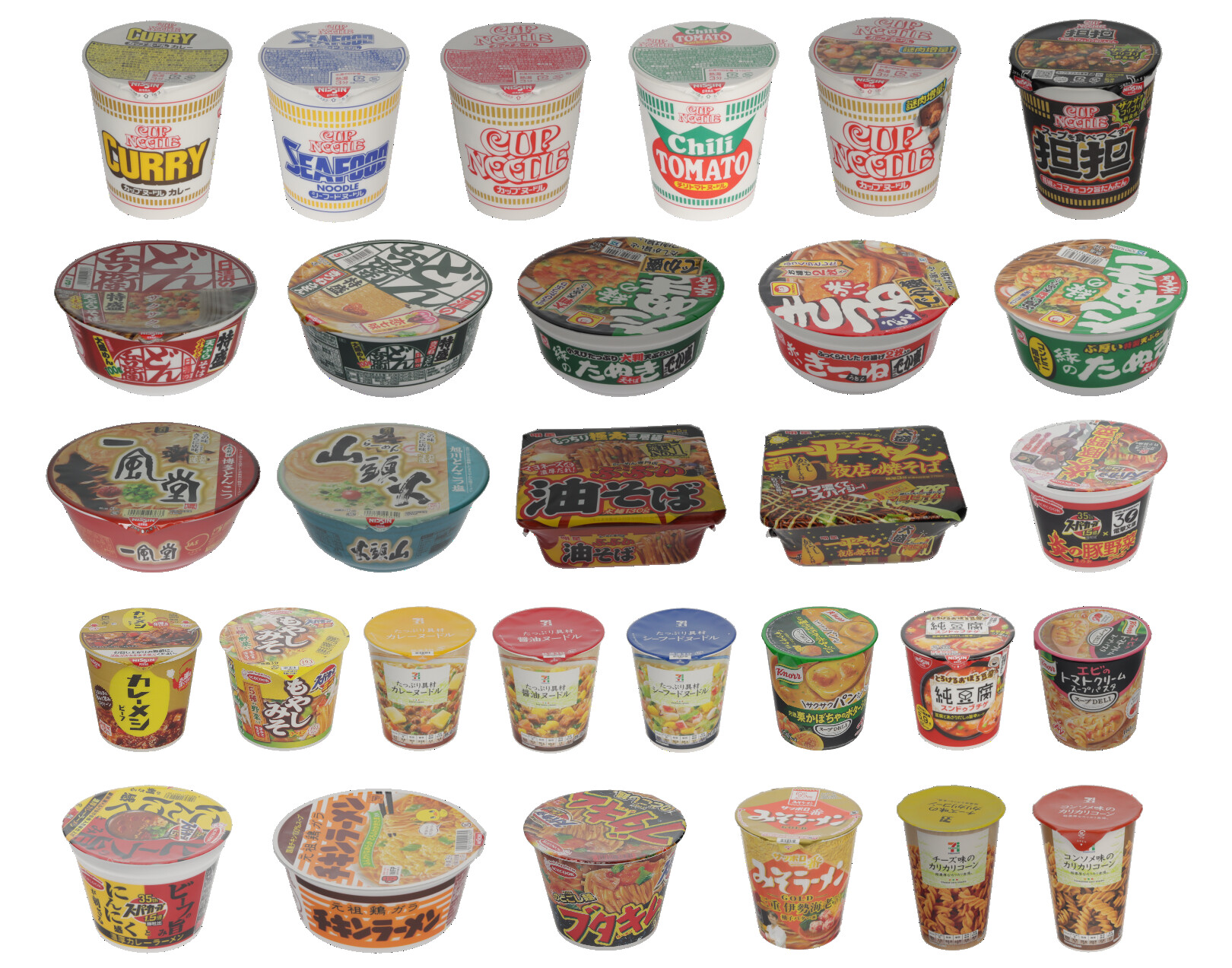}
	\caption{30 Objects recorded for our \emph{Ramen} dataset of common Japanese cup noodles available at most supermarkets.}
	\label{fig:konbini_dataset}
\end{figure}

\subsection{Data Set}
\label{subsec:konbini_dataset}

The focus of our research is on the development of robotic systems in the service sector to support personnel in retail. 
With Japan having one of the highest population densities, it boasts a considerable number of 24-hour convenience stores known as \emph{konbini}~\cite{numkonbinis}.
Introducing robots into these compact retail spaces can prove to be a valuable asset for tasks like restocking products and efficiently managing inventory. 
Therefore we focused on the specific product group of ramen noodles gathered in the \emph{Ramen} dataset.
The dataset comprises of 30 varieties of cup noodles, readily available in most mini markets. A comprehensive summary of all products is illustrated in Fig. \ref{fig:konbini_dataset}.


For recording the objects, we employed the \emph{3D PhotoBench 280} from \emph{Ortery} \cite{Ortery3DPhotoBench280}, a commercial $360^\circ$ turntable system, alongside the \emph{3D MultiArm 2000} \cite{Ortery3DMultiArm2000} camera system. This setup enabled synchronous image acquisition from five different angles capturing 150 images for each hemisphere and applying automatic background removal. The process is heavily inspired by \emph{Neural Scanning}~\cite{neuralscanning}.

The initial step involved scanning a planar calibration board with a feature-rich surface and an ArUco marker of known size. 
This process, aided by COLMAP~\cite{schoenberger2016sfm}, allowed for precise pose computation. 
Subsequently, the scene (including poses and point cloud) was scaled to achieve a metric reconstruction~\cite{CherryPicker} and aligned to the $xy$ plane ensuring the normal vector of the visible plane faced the positive $\mathbf{z}$-direction.
The poses obtained from the board were repurposed as a calibration reconstruction.
For each cup noodle product, the poses from the upper hemisphere's calibration reconstruction were utilized. 
The sparse point cloud was then recomputed by triangulating the matched feature points from the product images.
To capture photometric information about the bottom part, we scanned the bottom hemisphere of the flipped product symmetrically. 
However, it was not possible to reuse the calibration target for the bottom hemisphere, as the flipping of the object altered the pose locations. 
Our methodology involved registering the bottom images into the existing top reconstruction, resulting in approximately 270 registered images per product.
Each object was reconstructed using 3DGS and a low-poly mesh was extracted from the point cloud (as detailed in section \ref{subsec:gs_object}). 
To integrate these meshes into the physics engine, we store the information for every object, including environment objects, in a Unified Robotics Description Format (URDF) file, documenting both visual and collision model information.

In addition to the \emph{Ramen} dataset, we provide \emph{PEGASET} as a second dataset. 
It comprises a selection of 21 objects from the well-known YCB-V Dataset \cite{PoseCNN}. 
By including these fundamental objects, we aim to encourage broader adoption by the robotics community.

\section{Experiments}
This section presents experiments demonstrating the successful use of Universal Robots \textit{UR5} to execute real-world pick-and-place operations on data generated by \emph{PEGASUS}.

We selected the Deep Object Pose (DOPE) \cite{DOPE} network structure for our experiments. 
For dataset generation, three distinct data sets were created, each comprising 60,000 images featuring a single cup noodle, set in three different environments. 
In total, we generated 2,000 unique scenes with 30 images per scene, captured from various perspectives. 
The generation process took 6 hours on a laptop with an Intel i9 12th Gen CPU and NVIDIA RTX 3080 Ti (Mobile) GPU.
Regarding training, we utilized the default hyper-parameters of DOPE and trained the network for 15 epochs. 
The \textit{UR5} was configured to accurately pick the center of the cup noodle and place it into a basket. 
Our experiments successfully demonstrated the capability of the robot to sequentially pick up 10 out of 10 cup noodles in a row, as well as to grasp various types of cup noodles.

\section{Limitations}

Our method, while effective, is not without its limitations. 
One significant shortfall is the absence of realistic shadow rendering in our system. 
Consequently, incorporating shadow maps or screen space ambient occlusion represents a natural and necessary next step in our development process. 
Additionally, when placing objects within an environment, our current approach does not account for re-lighting, scattering, refraction, or reflection. 
This omission can result in scenes that appear somewhat unnatural.
Another challenge we faced involves scanning texture-less environments, which often leads to a noisy Gaussian splatting reconstruction. 
This noise manifests as large Gaussian splats that may overlap or interfere with objects, potentially causing visual artifacts. 
Addressing these issues is crucial for enhancing the realism and visual fidelity of our rendered scenes.

\section{Conclusion}

We have introduced \emph{PEGASUS}, a versatile dataset generator designed to enhance accuracy and quality in object pose estimation. Alongside \emph{PEGASUS}, we present the \emph{Ramen} dataset, which includes over 30 diverse products. 
The dataset generator adeptly creates photorealistic renderings, semantic masks, and depth maps, and captures the object pose.
\emph{PEGASUS} is specifically engineered to generate domain-specific data sets, aiding in the fine-tuning of neural networks that extend beyond mere pose estimation tasks.

To further empower \emph{PEGASUS}, it is crucial to accumulate a more extensive collection of environments and objects. 
This expansion is key to evolving toward a more generalizable dataset generator. Another intriguing avenue for enhancement involves applying augmentations directly to the objects or their environments. 
Techniques like diffusion models, such as \emph{GaussianDreamer}~\cite{gaussiandreamer} or \emph{Rosie}~\cite{rosie}, can be employed to alter the shape and appearance of objects and environments, offering a new dimension of flexibility in dataset generation.

Exploring the scanning of more complex scenes using LIDAR-based 3DGS presents an exciting opportunity. This approach could significantly enhance the realism of the environments integrated into our system. By leveraging LIDAR's detailed spatial data, we can capture intricate scene details and textures, paving the way for even more lifelike and accurate representations in our dataset generation process.

\section*{Acknowledgment}
We extend our sincere gratitude to \textbf{Abdullah Mustafa} for his valuable feedback and to \textbf{Toshio Ueshiba} for his extensive expertise in UR5.


\begin{thebibliography}{99}






\bibitem{PoseCNN}
Y. Xiang \emph{et al.},
"PoseCNN: A Convolutional Neural Network for 6D Object Pose Estimation in Cluttered Scenes,"
\emph{CoRR}, 2017.

\bibitem{HOPE}
S. Tyree \emph{et al.},
"6-DoF Pose Estimation of Household Objects for Robotic Manipulation: An Accessible Dataset and Benchmark," \emph{IROS}, 2022.

\bibitem{LINEMOD}
S. Hinterstoisser \emph{et al.},
"Model Based Training, Detection and Pose Estimation of Texture-Less 3D Objects in Heavily Cluttered Scenes,"
\emph{ACCV}, 2012.

\bibitem{FAT}
J. Tremblay \emph{et al.},
"Falling Things: A Synthetic Dataset for 3D Object Detection and Pose Estimation,"
\emph{CoRR}, 2018.


\bibitem{NDDS}
T. To \emph{et al.},
"NDDS: NVIDIA Deep Learning Dataset Synthesizer," 2018.

\bibitem{blenderproc2}
M. Denninger \emph{et al.},
"BlenderProc2: A Procedural Pipeline for Photorealistic Rendering,"
\emph{Journal of Open Source Software}, 2023.

\bibitem{nerf}
B. Mildenhall \emph{et al.}, "NeRF: Representing Scenes as Neural Radiance Fields for View Synthesis,"
\textit{ECCV}, 2020.

\bibitem{nerf_survey}
K. Gao, \emph{et al.}, "NeRF: Neural Radiance Field in 3D Vision, A Comprehensive Review," ArXiv, 2023

\bibitem{GaussianSplatting}
B. Kerbl \emph{et al.}, "3D Gaussian Splatting for Real-Time Radiance Field Rendering," \emph{SIGGRAPH}, 2023.

\bibitem{BOP}
M. Hodan \emph{et al.}, "BOP: Benchmark for 6D Object Pose Estimation," \emph{ECCV}, 2018.

\bibitem{DOPE}
J. Tremblay \emph{et al.}, "Deep Object Pose Estimation for Semantic Robotic Grasping of Household Objects," \emph{CoRL}, 2018.

\bibitem{point_based_rendering_1}
  G. Kopanas \emph{et al.}, "Point-Based Neural Rendering with Per-View Optimization,"
  \emph{Computer Graphics Forum}, 2021.

\bibitem{point_based_rendering_2}
  G. Kopanas \emph{et al.},
  "Neural Point Catacaustics for Novel-View Synthesis of Reflections,"
  \emph{ACM Transactions on Graphics}, 2022.

\bibitem{yang2023deformable3dgs}
Z. Yang \emph{et al.}, "Deformable 3D Gaussians for High-Fidelity Monocular Dynamic Scene Reconstruction," \emph{ArXiv}, 2023.

\bibitem{EWA}
M. Zwicker \emph{et al.}, "EWA volume splatting," \emph{IEEE Visualization}, EWA volume splatting, 2001. 

\bibitem{rosie}
Tianhe Yu \emph{et al.}, "Scaling Robot Learning with Semantically Imagined Experience,"
\emph{ArXiv}, 2023.

\bibitem{gaussiandreamer}
Taoran Yi \emph{et al.}, "GaussianDreamer: Fast Generation from Text to 3D Gaussian Splatting with Point Cloud Priors," \emph{ArXiv}, 2023.

\bibitem{schoenberger2016sfm}
J. L. Schönberger and J. M. Frahm, "Structure-from-Motion Revisited," CVPR, 2016.

\bibitem{instantNGP}
T. M\"uller \emph{et al.}, "Instant Neural Graphics Primitives with a Multiresolution Hash Encoding,"
\emph{SIGGRAPH},  2022.

\bibitem{NeRFShop}
C. Jambon \emph{et al.}, "NeRFshop: Interactive Editing of Neural Radiance Fields". \emph{I3D}, 2023.

\bibitem{plenoctree}
  A. Yu \emph{et al.},
  "PlenOctrees for Real-time Rendering of Neural Radiance Fields,"
  \emph{CoRR}, 2021.

\bibitem{clipnerf}
C. Wang \emph{et al.}, "CLIP-NeRF: Text-and-Image Driven Manipulation of Neural Radiance Fields," CVPR, 2022.

\bibitem{CLIP}
A. Radford et al., "Learning Transferable Visual Models From Natural Language Supervision,"
\emph{ICML}, 2021.

\bibitem{ADOP}
D. Rückert, L. Franke and M. Stamminger.
\textit{ADOP: Approximate Differentiable One-Pixel Point Rendering}.
CoRR. 2021.


\bibitem{franke2024trips}
L. Franke \emph{et al.}, "TRIPS: Trilinear Point Splatting for Real-Time Radiance Field Rendering," \emph{Eurographics}, 2024.



\bibitem{neuralscanning}
F. Erich \emph{et al.},
"Neural Scanning: Rendering and Determining Geometry of Household Objects Using Neural Radiance Fields".
\emph{SII}, 2023.

\bibitem{neurallabeling}
F. Erich \emph{et al.},
"NeuralLabeling: A versatile toolset for labeling vision datasets using Neural Radiance Fields," \emph{ArXiv}, 2023.

\bibitem{hodan2019photorealistic}
T. Hodan \emph{et al.},
"Photorealistic Image Synthesis for Object Instance Detection".
\emph{ICIP}, 2019.

\bibitem{pitteri2019object}
G. Pitteri \emph{et al.},
"On Object Symmetries and 6D Pose Estimation from Images".
\emph{3DV}, 2019.


\bibitem{CherryPicker}
L. Meyer, et al., 
"CherryPicker: Semantic Skeletonization and Topological Reconstruction of Cherry Trees,"
\emph{CVPRW}, 2023.

\bibitem{nealen2006laplacian}
A. Nealen \emph{et al.},
"Laplacian Mesh Optimization," \emph{GRAPHITE}, 2006.

\bibitem{YCB_Objects}
B. Calli \emph{et al.},
"Benchmarking in Manipulation Research: The YCB Object and Model Set and Benchmarking Protocols,"
\emph{IEEE Robotics and Automation Magazine}, 2015.

\bibitem{edelsbrunner1994threedimensional}
H. Edelsbrunner and E. Mücke, "Three-dimensional alpha shapes". \emph{ACM Transactions on Graphics}, 1994.

\bibitem{pyransac}
  L. Mariga,
  \emph{pyRANSAC-3D},
  2022,
  DOI: 10.5281/zenodo.7212567,.

\bibitem{GreenSH}
R. Green, "Spherical Harmonic Lighting: The Gritty Details," 2003.

\bibitem{JaroszAppendix}
W. Jarosz, "Efficient Monte Carlo Methods for Light Transport in Scattering Media,"
\emph{Dissertation}, 2008.

\bibitem{Kautz}
Jan Kautz, Peter-Pike Sloan, and John Snyder. 2002. "Fast, arbitrary BRDF shading for low-frequency lighting using spherical harmonics," Eurographics, 2002.

\bibitem{e3nn}
M. Geiger et al., "Euclidean neural networks: e3nn," \emph{GitHub}, 2022.

\bibitem{pybullet}
E. Coumans and Y. Bai, "PyBullet, a Python module for physics simulation for games, robotics and machine learning," 2016-2021.


\bibitem{numkonbinis}
Statista, \textit{Number of convenience stores in Japan from 2013 to 2022}, 2023.

\bibitem{Ortery3DPhotoBench280}
  \emph{Ortery 3D PhotoBench 280}, Ortery,
  2023.

\bibitem{Ortery3DMultiArm2000}
\textit{3D MultiArm 2000}, Ortery, 2023.


\end{thebibliography}
\end{document}